# CNN-based Human Detection for UAVs in Search and Rescue

Nikite Mesvan

*Vel Tech - Technical University*
*Chennai, India*

**Abstract**

*The use of Unmanned Aerial Vehicles (UAVs) as a substitute for ordinary vehicles in applications of search and rescue is being studied all over the world due to its flexible mobility and less obstruction, including two main tasks: search and rescue. This paper proposes an approach for the first task of searching and detecting victims using a type of convolutional neural network technique, the Single Shot Detector (SSD) model, with the Quadcopter hardware platform, a type of UAVs. The model used in the research is a pre-trained model and is applied to test on a Raspberry Pi model B, which is attached on a Quadcopter, while a single camera is equipped at the bottom of the Quadcopter to look from above for search and detection. The Quadcopter in this research is a DIY hardware model that uses accelerometer and gyroscope sensors and ultrasonic sensor as the essential components for balancing control, however, these sensors are susceptible to noise caused by the driving forces on the model, such as the vibration of the motors, therefore, the issues about the PID controller, noise processing for the sensors are also mentioned in the paper. Experimental results proved that the Quadcopter is able to stably flight and the SSD model works well on the Raspberry Pi model B with a processing speed of 3 fps and produces the best detection results at the distance of 1 to 20 meters to objects.*

***Keywords:*** *human detection; unmanned aerial vehicles (UAVs); quadcopter; convolutional neural network; Single Shot Detector (SSD);*

## 1. INTRODUCTION

Nowadays, Unmanned Aerial Vehicles (UAVs) are increasingly being used by individuals and organizations because of its wide applications in commercial, military, medical, and especially in search and rescue efforts. What makes UAVs attractive to search and rescue operations is their ability to fly in and move out disaster-stricken areas and inaccessible locations that take pilots and crew to danger. Moreover, the cost of using UAVs is much smaller than helicopters or airplanes. Currently, drones are often equipped with cameras to view the scene in real time. Flights near the disaster areas and the images obtained could help researchers assess damage and find survivors. Furthermore, UAVs can provide medical supplies to humans in quarantine or inaccessible areas. These are tasks that experts say will take much time and manpower to complete with human power. Quadcopter, or drone, is a type of UAVs, which is composed of four motors, and also is the most common type of UAVs used in industry and research due to its flexibility in movement, compact size, and ease of production [1-7].

Image processing is a branch of digital signal processing. Image processing is increasingly proving its usefulness in many aspects, applied in many fields. At the same time, deep learning, a branch of artificial intelligence, is a set of techniques that use multiple layers of neural networks to solve many problems in the field of image recognition, speech recognition, and many more. The combination of deep learning and image processing can be said to be the brain and the eyes of machines and is gradually replacing the human in the works related to detection and identification of objects. Productivity is faster and more accurate with the ability to process millions of images continuously with more speed and accuracy than the human can do [8-25].

Currently, the search and rescue mission is researched by experts, especially in supporting victims of a disaster. A potential approach proposed in this paper is to use drone as a supporting robot because its traveling path is less blocked from the air than that of the unmanned ground vehicles (UGVs) [26]. On the other side, one of the most important parts of the mission is human detection. Researches now on object detection are still developed day by day [27,28,29,30]. AlexNet, VGGNet, GoogLeNet are some of the pioneering deep network architectures. And now, YOLO and SSD are the state-of-the-art deep learning object detection methods that are demonstrating promising results and yielding faster detectors with impressive accuracy [8-25]. This paper presents a solution for human detection using a deep learning neural network technique, Single Shot Detector (SSD), with a Quadcopter platform in the application of search and rescue.

This paper is organized as follows. Section 2 firstly describes the quadcopter platform, then the PID controllers and a complementary filter for the quadcopter stabilization are presented as well. Next, an overview of convolutional neural network (CNN) technique and a state-of-the-art algorithm for object detection using a CNN technique, the Single Shot Detector (SSD), are discussed in section 3. Section 4 indicates the experimental results. Finally, the conclusions of the paper are presented in section 5.

## 2. THE HARDWARE PLATFORM AND CONTROL

### 2.1 Experimental Apparatus

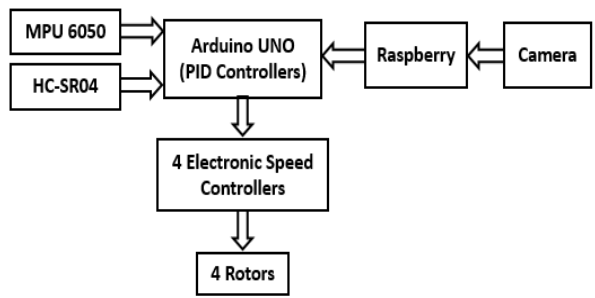

*Figure 1. The block diagram of the system.*

Figure 1 depicts the block diagram of the system while figure 2 shows the hardware platform used in this research.

The platform used in this research is a DIY hardware platform. S500 Quadcopter Frame Kit is increasingly used in education and research because of its low cost, robustness and easily assembled.

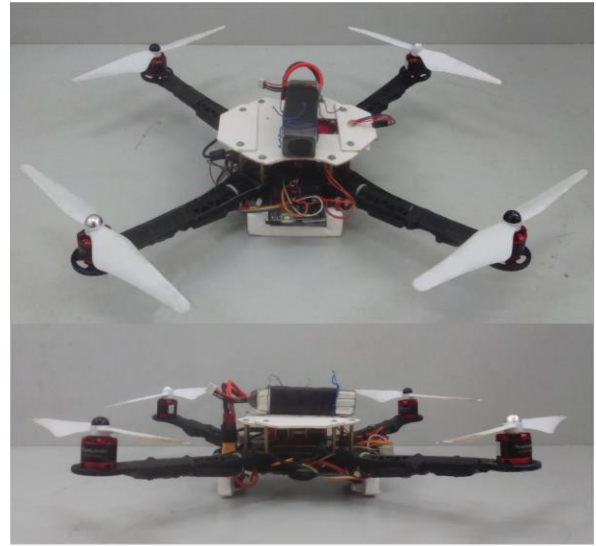

*Figure 2. The Quadcopter platform.*

This quadcopter, which is also a hardware platform used in [1], consists of a microcontroller Arduino Uno, 4 Electric Speed Controllers (ESCs), 4 Brushless DC motors, an Ultrasonic Sensor HC-HR04, a MPU 6050 sensor equipped with 6-degree-of-freedom inertial measurement unit: 3-axis gyroscope and 3-axis accelerometer, a Raspberry pi 3 model B, a 5.0 MP camera for the maximum resolution is 720p, and the resolution of 640x480 with 30fps at the bottom to look down vertically, and a 3800mAh battery for continuous flights from 10 to 15 minutes. The quadcopter can achieve the speed of about 3 m/s and operate very well for both indoor and outdoor.

### 2.2 PID Controller

PID (Proportional-Integral-Derivative) controller is the commonest control algorithm used in many applications to optimize the system automatically. A PID controller is capable of controlling the system to meet the quality criteria such as short transient time, fast response, and reducing the overshoot for the system [1-6]. Figure 3 shows the block diagram of a PID controller in a feedback loop.

In which, $r(t)$ is the desired process value or set-point (SP), $y(t)$ is the measured process value (PV), $e(t)$ is the error value which is the difference between the SP and the PV. The PID controller attempts to minimize the error over time by adjustment of a control variable $u(t)$.

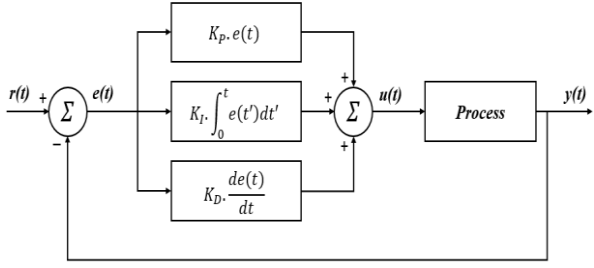

*Figure 3. PID controller.*

The control function can be expressed mathematically as follows [1-6]:

$$u(t) = K_P \cdot e(t) + K_I \cdot \int_0^t e(t')d(t') + K_D \cdot de(t)/dt \quad (1)$$

Next part presents PID controllers for the quadcopter balancing. The feedback are yaw, pitch and roll angle values calculated from the accelerometer sensor MPU 6050.

## 2.3 Quadcopter Dynamics

The yaw, pitch and roll angles of the quadcopter are initialized as figure 4.

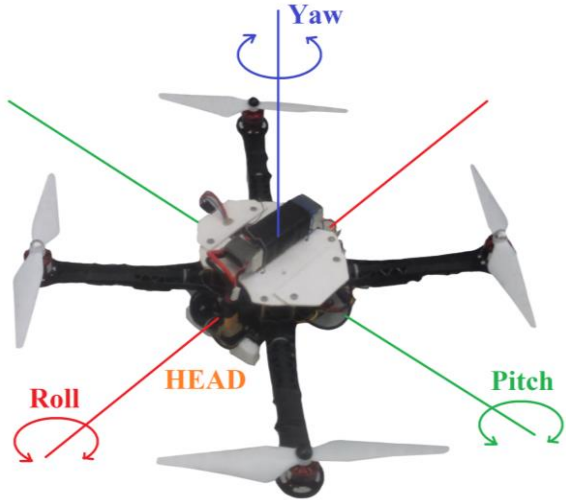

*Figure 4. The yaw, pitch and roll angles.*

Equation to calculate the angles from the gyroscope [1][3]:

$$\begin{bmatrix} \frac{\delta\psi}{\delta t} \\ \frac{\delta\theta}{\delta t} \\ \frac{\delta\phi}{\delta t} \end{bmatrix} = \frac{1}{\cos\phi} \times \begin{bmatrix} 0 & \sin\phi & \cos\phi \\ 0 & \cos\phi\cos\theta & -\sin\phi\cos\theta \\ \cos\theta & \sin\phi\sin\theta & \cos\phi\cos\theta \end{bmatrix} \times \begin{bmatrix} \omega_x \\ \omega_y \\ \omega_z \end{bmatrix} \quad (2)$$

Where $\Psi$, $\theta$, $\Phi$ are the angles at the previous time (t-1). So, the real values of them at the current time are computed as below:

$$\begin{bmatrix} \psi(t) \\ \theta(t) \\ \phi(t) \end{bmatrix} = \begin{bmatrix} \psi(t-1) \\ \theta(t-1) \\ \phi(t-1) \end{bmatrix} + \begin{bmatrix} \frac{\delta\psi}{\delta t} \\ \frac{\delta\theta}{\delta t} \\ \frac{\delta\phi}{\delta t} \end{bmatrix} \times \Delta t \quad (3)$$

Based on x-axis, y-axis and z-axis accelerations measured from the accelerometer, the roll and pitch angles are computed [1][3]:

$$\begin{bmatrix} A_x \\ A_y \\ A_z \end{bmatrix} = \begin{bmatrix} \cos\theta\cos\psi & \cos\theta\cos\psi & -\sin\theta \\ \cos\psi\sin\theta\sin\phi & \cos\phi\cos\psi + \sin\theta\sin\phi\sin\psi & \sin\theta\cos\phi \\ \cos\psi\sin\theta\cos\phi & -\sin\phi\cos\psi + \sin\theta\cos\phi\sin\psi & \cos\theta\cos\phi \end{bmatrix} \times \begin{bmatrix} 0 \\ 0 \\ 1 \end{bmatrix} \quad (4)$$

So, the actual roll and pitch angles value from the accelerometer are calculated as below:

$$Roll = \phi = \arcsin\left(\frac{A_y}{\cos\theta}\right) \quad (5)$$

$$Pitch = \theta = \arcsin(A_x) \quad (6)$$

In addition, in this platform, an ultrasonic sensor is used to maintain the altitude of the quadcopter in space. In this case, the echo time signal from the ultrasonic sensor is fed back to a PID controller [1].

Ultimately, apply the yaw, pitch, roll angles and the echo time signal to the PID controllers. The system finally is structured as figure 5:

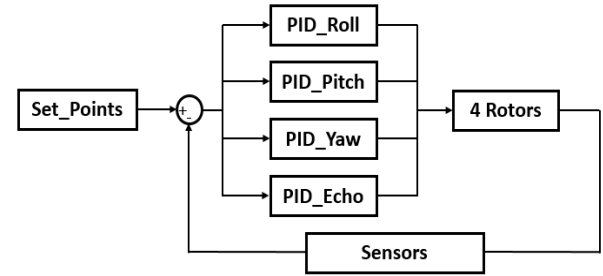

*Figure 5. The overall control structure of the quadcopter.*

## 2.4 Noise Processing

An accelerometer itself is a sensor susceptible to interferences. In quadcopters, the driving forces of the system, such as the vibration of the motors, will also affect the accelerometer. Therefore, using a filter is definitely necessary [1-2][7].

Besides, because of the integration over time, the measurement of the gyroscope tends to drift and do not return to zero when the system returns to its original position. The gyro data is only reliable in a short term and it starts to drift in a long term.

A complementary filter indicates an efficient solution. In the short term, the data from the gyroscope (*gyrData*) is used because of its high accuracy. In the long term, the data from the accelerometer (*accData*) is used since it does not drift. In the simplest form, the filter formula looks as below [1-2][7]:

angle = alpha(angle+gyrData.dt)+(1-alpha)accData (7)

In the equation (6), *alpha* is the filter coefficient (0<*alpha*<1) and *angle* is the output of the filter. Roll and pitch angle values are updated every iteration in a infinite loop. The filter will check whenever the values measured from the accelerometer is either reasonable or not. If any value is too large or too small, it is a complete disturbance, and the complementary filter attempts to decrease the influence of this disturbance for the better computation. Figure 6 shows the original signal of roll angle measured from the accelerometer over time (blue) and the processed signal with a complementary filter (red). It is facile to realize that the blue one is easily affected by noise and the red one is a better input for the controller [1].

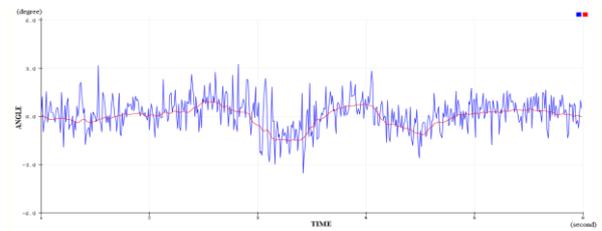

*Figure 6. Complementary filter.*

## 3. SINGLE SHOT DETECTOR (SSD)

### 3.1 Convolutional Neural Network

Convolutional Neural Network (CNN, or ConvNet) is a state-of-the-art technique at the current time that helps build intelligent systems with high accuracy based on the artificial neural network [8-17].

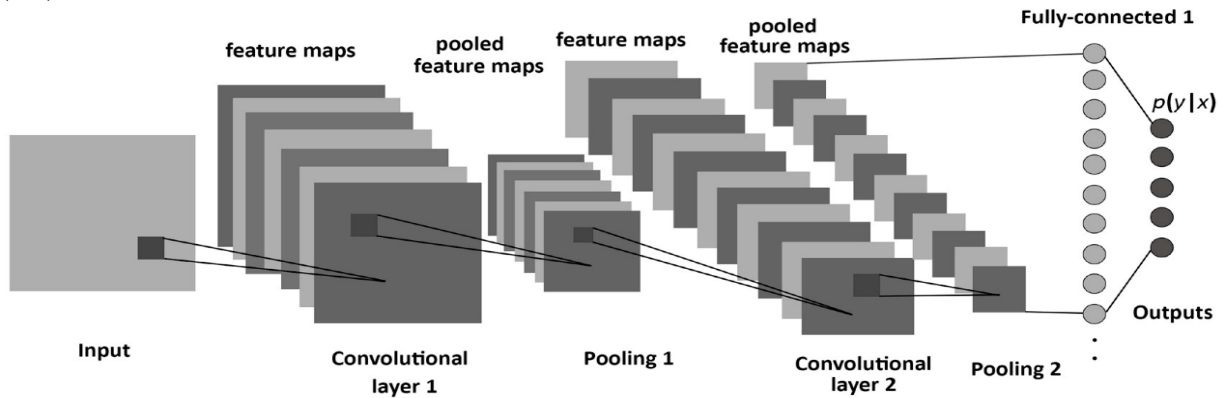

*Figure 7. Architecture of a CNN.*

A ConvNet is a sequence of layers, consists of an input and an output layer, as well as multiple hidden layers. The hidden layers commonly include three main types of layers: convolutional layers, pooling layers, and fully connected layers (exactly as seen in regular Neural Networks). For instance, these layers form a full ConvNet architecture as figure 7. In more detail:

Input layer: holding the values of image pixels, images may be binary images or three color channels images (RGB images).

Convolutional layer: using convolution, a mathematical operation, to calculate the output of neurons that are connected to the input or the previous convolutional layers.

ReLU (Rectified Linear Unit) layer: applying an activation function generally after convolutional layer, such as the max(0,x) with a threshold at 0.

Pooling layer: performing a down-sampling operation along the spatial dimensions (width and height).

Fully connected layer: as ordinary neural networks, each neuron in this layers will be connected to all the numbers in the previous volume, and the ultimate layer in this layers computes the scores of classes by commonly using the sigmoid function.

In this way, ConvNet transforms an original image from the original pixel values to the final class scores. The parameters in the convolutional layers and the fully connected layer will be trained with gradient descent method to minimize the loss function so that the class scores computed are consistent with the labels in the training data [8-17].

Object detection is a long-standing problem, yet in recent years, the rapid development of computer science and deep learning has drawn more attention to this issue. Nowadays, many deep learning algorithms have been deployed in a variety of forms and have been applied in various applications, such as face recognition, voice recognition, handwriting recognition, human detection, vehicles detection, etc [12-24].

There are various methods for object detection using Conventional Neural Network Techniques such as RCNN, Faster-RCNN, YOLO, SSD, etc [12-19]. While RCNN and Faster-RCNN demonstrate high-accuracy detecting results yet slow processing times, YOLO performs faster, however, with less accuracy. Single Shot Detector (SSD) achieves a good balance between speed and accuracy [12]; and also runs a good performance on smartphone and Raspberry Pi platforms. Hence, SSD is the model which is proposed to implement in the research. Table 1 depicts a speed and accuracy comparison of object detection algorithms based on VGG-16 architecture [13], with the input size 300x300, SSD produced an impressive results with 46 FPS and 74.3 mAP (results on Pascal VOC2007 test) [12].

**Table 1.** VGG16-based object detection algorithms: Speed and accuracy comparison [12].

| Method | mAP | FPS | Input resolution |
|---|---|---|---|
| Faster R-CNN | 73.2 | 7 | 1000x600 |
| YOLO | 66.4 | 21 | 480x480 |
| **SSD300** | **74.3** | **46** | **300x300** |

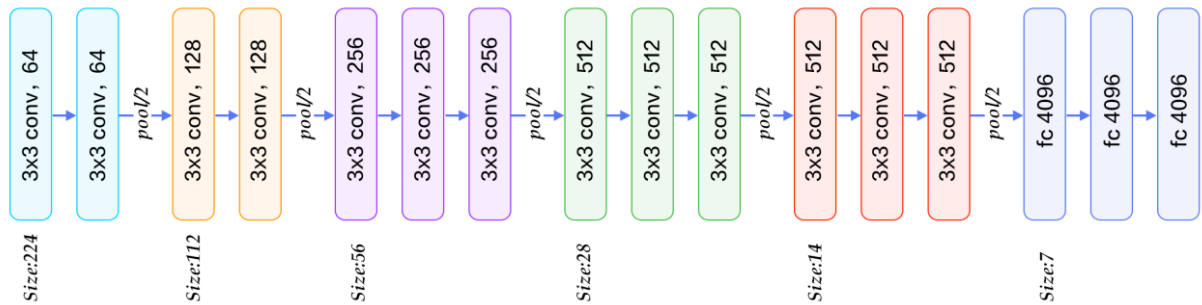

*Figure 8. The VGG-16 architecture.*

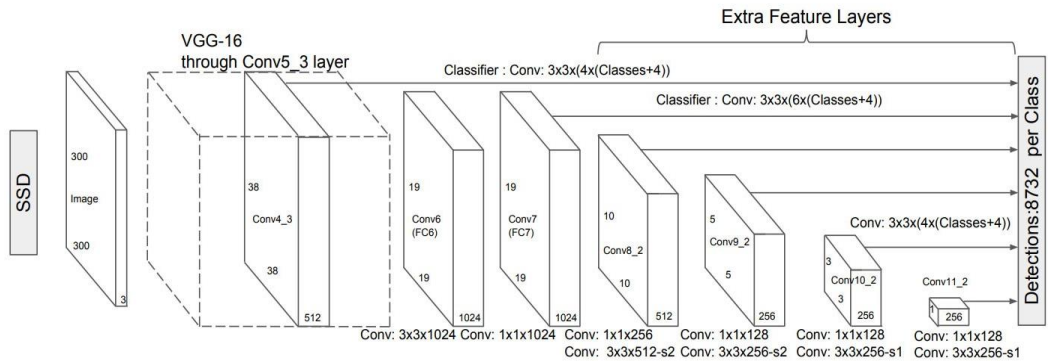

*Figure 9. The SSD architecture.*

### 3.2 The Single Shot Multibox Detector (SSD)

The SSD approach, which is based on a feed-forward neural network, produces a set of boxes in different sizes and scores for the confidence of the object appearance in those boxes, followed by the non-max suppression method to perform the final detection [12].

As depicted in figure 9, the SSD architecture is built based on the VGG-16 [13] network architecture (as figure 8) but removed the fully-connected layers [12]

The reason that VGG-16 is used as a base network due to its powerful performance [13] in high-quality image classification tasks and its popularity in transfer learning. Instead of fully-connected layers in the original VGG-16 architecture, a set of auxiliary convolutional layers is added from the Conv6 layer allowing the extraction of features on multiple scales and the size of every Conv layer is decreased gradually [12].

SSD only runs a convolutional neural network on the input image once and computes a feature map. First, a small $3 \times 3$ size kernel runs on this feature map to predict the bounding boxes and also produces the probability for every box. Moreover, similar to Faster-RCNN [17], SSD also uses anchor boxes [12][15][17-19] at different ratio and sizes to handle the problem of detecting objects in different scales. Because of the operation of the convolutional layers at different scales, it is able to detect objects of various scales.

As figure 10a, SSD needs input images and ground-truth boxes for each object during the training time. SSD divides an image into grid cells and evaluates a small set of anchor boxes with different ratios and sizes at each location in several different feature maps, for instance, 8×8 map and 4×4 map in figure 10b and 10c. For each anchor box, SSD predicts both shape and confidence for all objects [12].

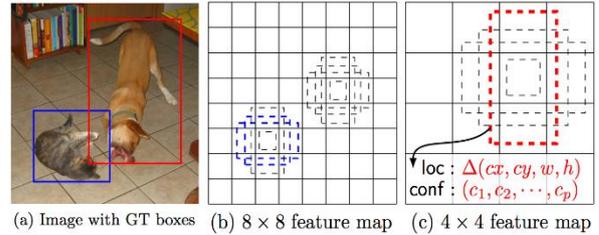

*Figure 10. Anchor boxes.*

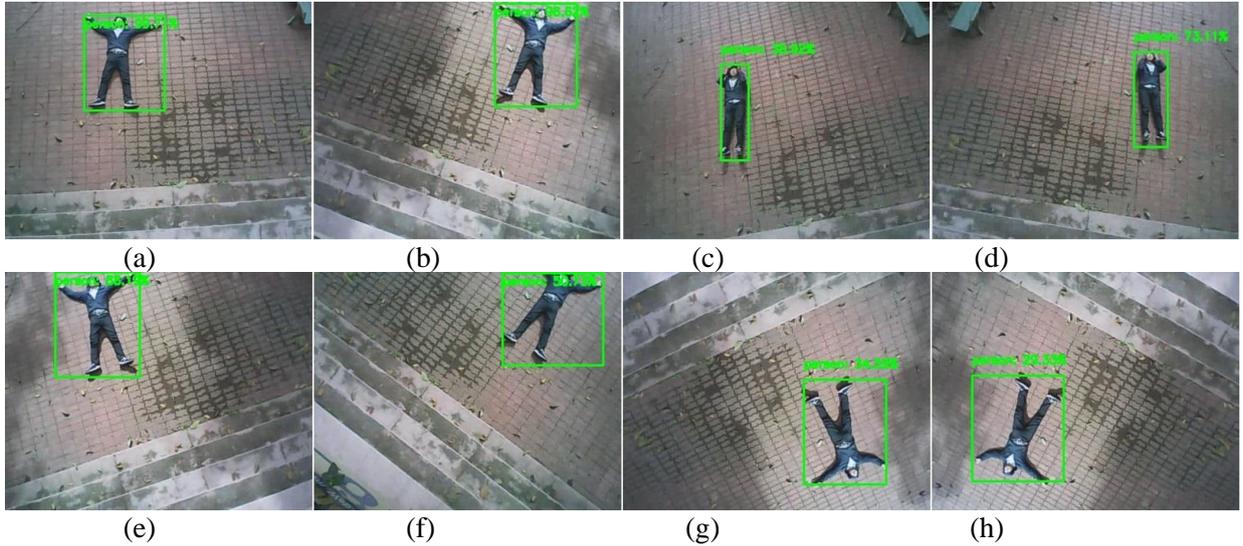

(a)     (b)     (c)     (d)

(e)     (f)     (g)     (h)

*Figure 11. Human detections at different postures and perspectives.*

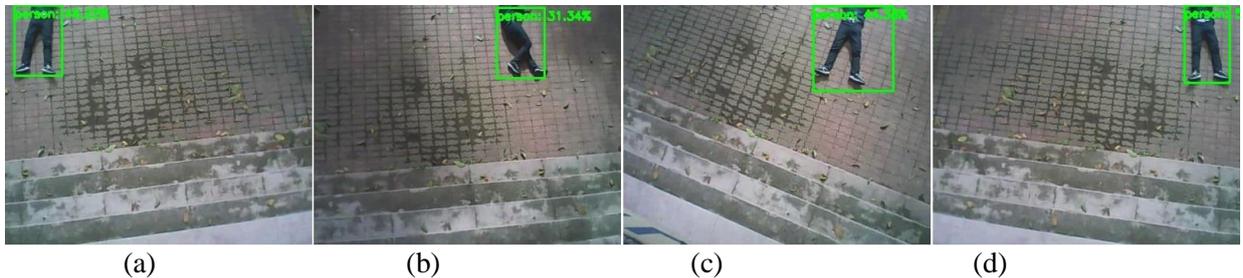

(a)     (b)     (c)     (d)

*Figure 12. Detection results in images which do not catch fully a person into the frame of the camera.*

The training mission is matching these anchor boxes with ground truth boxes. For example, in figure 10b-10c, with an 8x8 map, two anchors are identified as positive for the cat and the rests are negative, thus the network will be trained to match the positive ones with the ground-truth boxes. Similarly, in the 4x4 map, there is one anchor which is positive for the dog, the remaining anchors will be discarded. In figure 10c, *loc* defines the size of anchors, where $cx$, $cy$ are the (x,y) coordinates of the anchor's center in 2D image, and $w$, $h$ are the width and height of the anchor; *conf* corresponds to the confidence of each class [12][15][17-19].

## 4. EXPERIMENTAL RESULTS

The main objective of this section is to demonstrate the experimental results of the designed system and a comparison to previous methods.

The pre-trained model used in our research was trained on CAFFE [25], an open framework for deep learning, covering 21 categories: "background", "aero-plane", "bicycle", "bird", "boat", "bottle", "bus", "car", "cat", "chair", "cow", "dining-table", "dog", "horse", "motorbike", "person", "potted-plant", "sheep", "sofa", "train", "tv-monitor", for our research, the source code is adjusted to detect only "person".

Figure 11a-11h depict the experiments of human detection with the pre-trained model, the results proved that the model is able to detect people at different postures and perspectives. Table 2 indicates a speed comparison between SSD and previous researches using the Histogram of Oriented Gradients (HOG) Detector [20], the Haar Cascade method [21],

which were applied to human-searching UAVs, the experimental results demonstrated that the SSD model is better in human detection application. For further detail, SSD indicated a better performance than the HOG method and the Haar Cascade method with an average processing-speed of 3 fps (in the previous researches, the HOG Detector indicated an average computation time of 18.879 sec per image [20], and the Haar Cascade method produced a processing-speed of 1 fps [21]). At the same time, the accuracy of the human detection produced by the model is also evaluated that over 80% based on the experiments which are showed in figure 11 and figure 12.

**Table 2.** SSD: Speed comparison in human detection for drone with the previous methods (Haar Cascade and HOG).

| Method | FPS | Sec per image |
|---|---|---|
| HOG | 0.053 | 18.879 |
| Haar Cascade | 1 | 1 |
| **SSD** | **3** | **0.333** |

In more difficult contexts, described in figure 12a-12d, when images may not catch fully a person into the frame of the camera, the model still produces encouraging results, which previous methods such as Haar Cascade and HOG have not worked well [20-21].

## 5. CONCLUSIONS

This paper has proposed an approach for human search mission using a convolutional neural network technique, the Single Shot Detector (SSD) model, for UAVs in search and rescue application. The SSD model used in the research is a pre-trained model and also is suitable for running on the Raspberry Pi hardware platform; a single camera is attached at the bottom of a Quadcopter model for observation and detection. Experiments proved that the Quadcopter is able to stably flight, at the same time, the model works well on the Raspberry Pi with the processing speed of average 3 fps and performs better results at the distance of 1-20m from the camera to objects. In more difficult tasks, such as the images which do not display completely and fully a person into the frame of the camera, the model also demonstrates encouraging results.

Future works include building and training a better SSD model on a more powerful hardware platform for a better accuracy and a faster processing speed, and improving the research to adding localization function.